\documentclass[sigconf]{acmart}
\AtBeginDocument{%
  }

\copyrightyear{2025}
\acmYear{2025}
\setcopyright{acmlicensed}\acmConference[SIGSPATIAL '25]{The 33rd ACM International Conference on Advances in Geographic Information Systems}{November 3--6, 2025}{Minneapolis, MN, USA}
\acmBooktitle{The 33rd ACM International Conference on Advances in Geographic Information Systems (SIGSPATIAL '25), November 3--6, 2025, Minneapolis, MN, USA}
\acmDOI{10.1145/3748636.3763212}
\acmISBN{979-8-4007-2086-4/2025/11}

\usepackage{makecell}
\usepackage{subcaption}
\usepackage{enumitem}
\begin{document}

%%
%% The "title" command has an optional parameter,
%% allowing the author to define a "short title" to be used in page headers.
\title{Classical Feature Embeddings Help in BERT-Based Human Mobility Prediction}

%%
%% The "author" command and its associated commands are used to define
%% the authors and their affiliations.
%% Of note is the shared affiliation of the first two authors, and the
%% "authornote" and "authornotemark" commands
%% used to denote shared contribution to the research.
\author{Yunzhi Liu}
\orcid{0009-0007-5224-3580}
\affiliation{%
  \institution{UNSW Sydney}
  \city{Sydney}
  \state{NSW}
  \country{Australia}
}
\email{yunzhi.liu@unswalumni.com}

% Author information
\author{Haokai Tan}
\orcid{0009-0002-2507-712X}
\affiliation{%
  \institution{UNSW Sydney}
  \city{Sydney}
  \state{NSW}
  \country{Australia}
}
\email{haokai.tan@student.unsw.edu.au}

\author{Rushi Kanjaria}
\orcid{0009-0000-3311-2699}
\affiliation{%
  \institution{UNSW Sydney}
  \city{Sydney}
  \state{NSW}
  \country{Australia}
}
\email{r.kanjaria@unswalumni.com}

\author{Lihuan Li}
\orcid{0009-0001-9044-8532}
\affiliation{%
  \institution{UNSW Sydney}
  \city{Sydney}
  \state{NSW}
  \country{Australia}
}
\email{lihuan.li@student.unsw.edu.au}

\author{Flora D. Salim}
\orcid{0000-0002-1237-1664}
\affiliation{%
  \institution{UNSW Sydney}
  \city{Sydney}
  \state{NSW}
  \country{Australia}
}
\email{flora.salim@unsw.edu.au}

%%
%% By default, the full list of authors will be used in the page
%% headers. Often, this list is too long, and will overlap
%% other information printed in the page headers. This command allows
%% the author to define a more concise list
%% of authors' names for this purpose.
\renewcommand{\shortauthors}{Liu et al.}

%%
%% The abstract is a short summary of the work to be presented in the
%% article.
\begin{abstract}
Human mobility forecasting is crucial for disaster relief, city planning, and public health. However, existing models either only model location sequences or include time information merely as auxiliary input, thereby failing to leverage the rich semantic context provided by points of interest (POIs). To address this, we enrich a BERT-based mobility model with derived temporal descriptors and POI embeddings to better capture the semantics underlying human movement. We propose STaBERT (Semantic-Temporal aware BERT), which integrates both POI and temporal information at each location to construct a unified, semantically enriched representation of mobility. Experimental results show that STaBERT significantly improves prediction accuracy: for single-city prediction, the GEO-BLEU score improved from 0.34 to 0.75; for multi-city prediction, from 0.34 to 0.56.
\end{abstract}

%%
%% The code below is generated by the tool at http://dl.acm.org/ccs.cfm.
%% Please copy and paste the code instead of the example below.
%%
% CCSXML Classification System


%%
%% Keywords. The author(s) should pick words that accurately describe
%% the work being presented. Separate the keywords with commas.

%\received{20 February 2007}
%\received[revised]{12 March 2009}
%\received[accepted]{5 June 2009}

%%
%% This command processes the author and affiliation and title
%% information and builds the first part of the formatted document.
\maketitle

\section{Introduction}
\label{chap:introduction}
Human mobility forecasting has become a significant research field with broad applications in disaster relief~\cite{Amiri2024} and city planning~\cite{Ratti2006}. Forecasting and understanding how people travel through and between urban places has become increasingly critical to addressing societal issues as cities grow more connected and complex~\cite{wang2019,kim2018}. Existing competitive methods~\cite{Terashima2023,Solatorio2023,mobb_koyama23} for human mobility prediction employ Transformers~\cite{vaswani2017} to capture the spatiotemporal dependencies of daily human movement patterns. However, they either only model location sequences or include time information merely as auxiliary input, thereby failing to leverage the rich semantic context provided by points of interest (POIs). Only one method~\cite{Suzuki2023} has attempted to incorporate POI-related features—specifically, the frequency of visits to different POI categories—into a traditional Support Vector Regression (SVR) model. However, this approach lacks the modeling capacity of modern deep learning architectures and performs significantly worse than Transformer-based models.

In this paper, we propose a Semantic-Temporal aware BERT model (STaBERT), which integrates both derived temporal descriptors and POI embeddings at each visited location to learn a unified representation of human mobility, yielding a more comprehensive and semantically enriched model of movement patterns. The results show that the inclusion of these two types of features significantly improves prediction accuracy. For single-city prediction, the GEO-BLEU score increased from 0.34 to 0.75; for multi-city prediction, it rose from 0.34 to 0.56.

\section{Related Work}
\label{chap:literature_review}
According to the winner of HuMob Challenge 2023 H. Terashima et al.~\cite{Terashima2023}, developed a model based on BERT~\cite{Devlin2018} and extended it with a Transformer. Spatio-Temporal BERT~\cite{Meisaku2024}, which won the 2024 HuMob Challenge, uses LP-BERT along with LSTM and ensembling strategies to capture temporal patterns across cities, but it does not explicitly incorporate POI information. Furthermore, according to another top team of HuMob Challenge 2023 M. Suzuki et al.~\cite {Masahiro2023},  they add POI and POI clustering into the SVR model.  What’s more, according to R. Jiang et al.~\cite {Jiang2021}, they indicate by combining human mobility data and city POI data, a more effective representation of human mobility can be expected. Unlike previous methods, our approach integrates both fine-grained temporal descriptors and POI embeddings directly into a Transformer-based architecture, enabling unified semantic-temporal modeling within a deep learning framework.

\section{Dataset Description and Problem Definition}
\label{sec:Dataset Description and Problem Definition}
The dataset for the single-city task comes from the HuMob Challenge 2023. The challenge is set in a dense Japanese metropolitan area, divided into 500m × 500m grid cells across a 200 × 200 map. The dataset records the movements of 100,000 individuals over 90 days at 30-minute intervals. Task 1 involves predicting the movements of 20,000 individuals for days 60–74 using partial data, while Task 2 focuses on predicting masked locations for 2,500 out of 25,000 individuals. Optional POI data (85-dimensional vectors) for each cell is provided as additional context.

The dataset for the multi-city task also comes from the HuMob Challenge 2024. The dataset structure is the same as that of the 2023 challenge. The task is to predict the movements of 3,000 individuals in each of cities B, C, and D during days 61–75, using movement data from city A (full trajectories of 100,000 individuals from day 1 to 75) as well as partial data from cities B, C, and D, which include a total of 25,000, 20,000, and 6,000 individuals, respectively.

\section{Methodology}
\label{chap:methodology}
\subsection{Semantic-Temporal aware BERT model (STaBERT)}
\label{sec:lpbert_poi_time_methodology}

\subsubsection{Workflow}

% \begin{comment}
% \begin{figure}[htbp]
%     \centering
%     \includegraphics[width=1\linewidth]{Workflow.png} % Assuming image file is named this
%     \caption{Workflow of STaBERT}
%     \label{fig:Workflow_Yunzhi} % Changed label to be unique
% \end{figure}

% The workflow of this specific sub-method is shown in Figure \ref{fig:Workflow_Yunzhi}. 
% \end{comment}
Initially, data from the 2023 and 2024 HuMob competitions is utilized. The LP-BERT model, a winning solution from the 2023 challenge, serves as the base model. POI data is incorporated by constructing POI embeddings and applying mean pooling to integrate these features. The model is first trained on the 2023 competition data to predict single-city mobility. Subsequently, temporal descriptors are introduced, and their impact is evaluated. In parallel, city embeddings are explored using the POI-enhanced model on the 2024 multi-city dataset. Finally, evaluation metrics across all model variants are summarized to assess both predictive accuracy and computational efficiency.

\subsubsection{Model description}
\begin{figure}[htbp]
    \centering
    \includegraphics[width=1\linewidth]{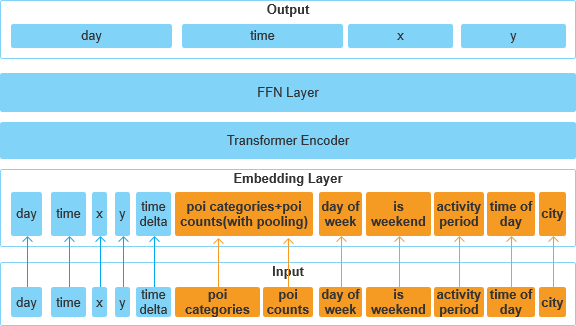} % Assuming image file is named this
    \caption{STaBERT model structure. Enhancements highlighted in orange}
    \label{fig:LP-BERT_with_POI_time_Yunzhi} % Changed label to be unique
\end{figure}

This model builds upon the original LP-BERT model~\cite{Li23} by incorporating POI information and detailed time descriptions. The architecture is illustrated in Figure \ref{fig:LP-BERT_with_POI_time_Yunzhi}, with improvements highlighted in orange. Input embeddings process features such as date, time, Location ID, time difference from the previous movement, POI categories/counts, day of the week, an indicator for weekends, activity period, and time of day. These embeddings are summed. For POI features, mean pooling is employed to derive a single, fixed-length representation per sequence or location. During training, Location IDs are randomly masked for $\alpha$ consecutive days, while other inputs remain visible. For efficiency, locations are predicted in parallel. A penalty factor $\beta$ is used during training to discourage repeated predictions of the same location on the same day. The original HuMob datasets use pseudo-identifiers for users, assigned sequentially from 0, which cannot be linked to any real individual. In our experiments, we exclude these pseudo-IDs from model inputs to maintain focus on general spatiotemporal patterns and to further ensure user-level privacy.

\subsubsection{Input of POI (Semantic features)}
POI information is crucial as it often reflects the purpose of visiting a particular location. For instance, knowing supermarket locations can enhance prediction accuracy for users who typically shop after work. A key challenge is fusing the multi-dimensional POI data effectively; mean pooling is employed for this purpose.

A dedicated POI representation module, featuring two embedding layers and a mean pooling operation, generates location-level embeddings based on POI category and count:
\begin{enumerate}[noitemsep, topsep=0pt]
    \item \textbf{Category-based POI Embedding:} Each POI category index $c_i \in \{0, \dots, 85\}$ is mapped to a dense vector $\mathbf{e}_i^{\text{cat}} = \text{Embedding}_{\text{cat}}(c_i) \in \mathbb{R}^d$ via a learned embedding lookup table.
    \item \textbf{Count-based POI Embedding:} Each POI count index $k_i \in \{0, \dots, 735\}$ is mapped to a dense vector \\ $\mathbf{e}_i^{\text{count}} = \text{Embedding}_{\text{count}}(k_i) \in \mathbb{R}^d$ using a separate learned embedding layer.
\end{enumerate}
Mean pooling is then applied across the embedded POIs to obtain a fixed-length representation for each location at each time step. Given POI embeddings $\mathbf{E} \in \mathbb{R}^{B \times T \times N \times d}$ (where $B$ is batch size, $T$ is sequence length, $N$ is the number of POIs per location, and $d$ is the embedding dimension) and a binary mask $\mathbf{M} \in \{0, 1\}^{B \times T \times N}$ indicating valid POIs, the masked mean embedding $\mathbf{\bar{E}}_{b,t}$ for batch $b$ and time step $t$ is computed as:
\[
\mathbf{\bar{E}}_{b,t} = \frac{\sum_{n=1}^{N} \mathbf{M}_{b,t,n} \cdot \mathbf{E}_{b,t,n}}{\max\left(1, \sum_{n=1}^{N} \mathbf{M}_{b,t,n}\right)} \in \mathbb{R}^d
\]

\subsubsection{Input of Time Description (Temporal Features)}
The day of the week and the specific time of day significantly influence travel patterns, distinguishing, for example, weekday routines from weekend activities, or typical travel times from late-night periods when movement is less common. The following time descriptions are incorporated: day of the week, an indicator for whether it is a weekend, the activity period, and the time of day. These are typically derived features:
% Corrected itemize environment - ensure it's not in math mode
\begin{itemize}[noitemsep, topsep=0pt]
  \item \texttt{day\_of\_week} = \( d \bmod 7 \) (where \( d \) is the day index).
  \item \texttt{is\_weekend} = 1 if \texttt{day\_of\_week} corresponds to Saturday or Sunday; 0 otherwise.
  \item \texttt{activity\_period}: Categorized based on the time slot \( t \) (e.g., “very\_early,” “early,” “active,” “high\_active,” “rest,” “deep\_rest”).
  \item \texttt{time\_of\_day}: Categorized, for example, as AM/PM based on the time slot \( t \).
\end{itemize}

These derived features are then embedded and integrated into the model’s input representation.

\section{Experiments}
\subsection{Evaluation Metrics}
We used GEO-BLEU and Dynamic Time Warping (DTW) to evaluate the accuracy of predictions.
GEO-BLEU~\cite{Shimizu2022} is a metric that emphasizes local features, similar to similarity measures used in natural language processing. A perfect match between predicted and actual results yields a score of 1.
DTW~\cite{dtw_wiki2025} is used to measure the temporal similarity between predicted and actual trajectories. Lower values indicate better alignment, with a score of 0 representing a perfect match.

\subsection{Base Model}

We use the winning model of the 2023 Human Mobility Prediction Challenge ~\cite{Terashima2023}, namely LP-Bert, as our base model. This model treats the trajectory prediction task as a BERT-based interpolation task, where the complete information of each location record is represented as a token and converted into a token embedding by a spatiotemporal embedding layer. 
The input of the model includes the day, time, x, y, and timedelta. Except for timedelta, all other inputs are raw data. Timedelta is the time difference from the previous movement.

\subsection{Experiment and Result}
We compare our model against the top three results from the 2023 and 2024 competitions.
The result for the single-city normal scenario is shown in Table \ref{Results for Single City Normal Scenario}.
The results show that our model significantly outperforms the top three entries from the 2023 competition. In particular, our GEO-BLEU scores are nearly twice as high as the baseline results.

\begin{table}[htbp]
    \centering
    \caption{Results for Single City Normal Scenario}
    \begin{tabular}{lcc} 
        \toprule 
        Model & GEO-BLUE $\uparrow$ & DTW $\downarrow$ \\
        \midrule 
        LP-Bert\cite{Terashima2023}(Winner of 2023) & 0.344 & 29.963 \\
        GeoFormer\cite{Solatorio2023}(2nd of 2023) & 0.316 & 26.216\\
        MOBB\cite{mobb_koyama23}(3rd of 2023) & 0.327 & 38.651 \\
        STaBERT & \underline{\textbf{0.750}} & \underline{\textbf{20.923}}\\
        \bottomrule 
    \end{tabular}
    \label{Results for Single City Normal Scenario}
\end{table}

% \begin{comment}
% 2. Add POI model (Add poi): Input includes day, time, x, y, timedelta, poi categories and poi counts.

% 3. Add POI and time description model (Add poi time): Input includes day, time, x, y, timedelta, poi categories, poi counts, day of week, is weekend, time of day, activity period

% % \begin{comment}
% Since the base model is a version reproduced from the 2023 champion's paper and is not an official code, the calculation results are different from those of the 2023 champion group. In order to reflect the competitiveness of the model, we add the results of the 2023 and 2024 champion as a control.

% Since efficiency is also one of the evaluation indicators of large models, we add Average training time per epoch (Time per epoch) as an efficiency evaluation indicator based on the accuracy evaluation indicator.
% \end{comment}

To assess the contribution of time description features and evaluate model efficiency, we conducted validation using half of the dataset. The results are presented in Table \ref{tab:Experiment 1}.

\begin{table}[htbp]
    \centering
    \caption{Results for Single City Normal Scenario: Smaller Training Split (Balancing Accuracy and Efficiency)}
    \begin{tabular}{lccc} 
        \toprule 
        Evaluation index & Base model & Add poi & Add poi time \\
        \midrule 
        GEO-BLEU $\uparrow$ & 0.247 & 0.354 & 0.496\\
        DTW $\downarrow$ & 64.972 & 49.944 & 34.778\\
        Time per epoch(min) $\downarrow$ & 12.6 & 37.0 & 37.5\\
        \bottomrule 
    \end{tabular}
    \label{tab:Experiment 1}
\end{table}

To evaluate the model’s performance in emergency scenarios, we conducted experiments using emergency scenario data. The results are shown in Table \ref{Experiment 3}.
It should be noted that this experiment was conducted without updating the POI data for the emergency scenario. Such untimely updates are common in emergency situations.

% \begin{comment}
% \begin{table}[htbp]
%     \centering
%     \caption{Results for Single City Emergency Scenario}
%     \begin{tabular}{lcc} 
%         \toprule 
%         Evaluation index & Add poi & Base model \\
%         \midrule 
%         GEO-BLEU & 0.264 & 0.224 \\
%         DTW & 52.399 & 44.774\\
%         Time per epoch(min) & 19.6 & -\\
%         \bottomrule 
%     \end{tabular}
%     \label{Experiment 3}
% \end{table}
% \end{comment}

\begin{table}[htbp]
    \centering
    \caption{Results for Single City Emergency Scenario}
    \begin{tabular}{lcc} 
        \toprule 
        Model & GEO-BLUE $\uparrow$ & DTW $\downarrow$ \\
        \midrule 
        LP-Bert\cite{Terashima2023}(Winner of 2023) & 0.224 & 44.774 \\
        GeoFormer\cite{Solatorio2023}(2nd of 2023) & 0.183 & \textbf{37.782}\\
        MOBB\cite{mobb_koyama23}(3rd of 2023) & 0.223 & 54.269 \\
        STaBERT & \underline{\textbf{0.264}} & \underline{52.399}\\
        \bottomrule 
    \end{tabular}
    \label{Experiment 3}
\end{table}

To evaluate the model’s performance on multiple cities, we applied it to datasets from multiple cities. The results are shown in Table \ref{Experiment 4}.

% \begin{comment}
% \begin{table}[htbp]
%     \centering
%     \caption{Results for Multi-city}
%     \begin{tabular}{lccc} 
%         \toprule 
%         Evaluation index & Add poi & Base model \\
%         \midrule 
%         GEO-BLEU & 0.563 & 0.319 \\
%         \bottomrule 
%     \end{tabular}
%     \label{Experiment 4}
% \end{table}
% \end{comment}

\begin{table}[htbp]
    \centering
    \caption{Results for Multi-city}
    \begin{tabular}{lcc} 
        \toprule 
        Model & GEO-BLUE $\uparrow$ & DTW $\downarrow$ \\
        \midrule 
        Multiple BERTs\cite{Multiple_BERTs}(Winner of 2024) & 0.319 & 28.21 \\
        Llama-3-8b\cite{Llama-3-8b}(2nd of 2024) & 0.309 & \textbf{27.96}\\
        Spatiotemporal BERT\cite{Meisaku2024}(3rd of 2024) & 0.305 & 30.45 \\
        STaBERT & \underline{\textbf{0.563}} & \underline{28.55}\\
        \bottomrule 
    \end{tabular}
    \label{Experiment 4}
\end{table}

% \begin{comment}
% In experiment 4, we tried to apply this model in multiple cities. We used Human mobility 2024 data. Due to time constraints, we only consider the data of cities B, C, and D. In line with the requirements of the competition, we predict the trajectories of the last 3,000 users from the 60th to the 74th day, and the rest of the data is used for training. We only apply the add poi model, and add the city name as input based on the model to establish city embedding to adapt to multiple city scenarios. The results are presented in Table \ref{Experiment 4}.    \looseness -1
% \end{comment}
From the above results, adding POI data has a great effect on improving the scenarios without special circumstances. It is not only effective for a single city, but also for multiple cities. Especially for large data sets, there is a very significant improvement. For scenarios with special circumstances, adding POI does not make a significant difference. Preliminary analysis shows that the possible reason is that the dataset does not reflect the status of POI points in special circumstances. For example, in special circumstances, they may be closed, which will have a significant impact on the prediction.

For time description, the accuracy is improved after adding the time description. Although the improvement in accuracy is not significant, adding time indicators does not significantly increase the training time, so the time description can be retained. In summary, adding POI and time description is the most optimal model solution.

\subsubsection{Analysis and visualization}
To further evaluate the impact of adding POI and time information on the model, we selected the normal dataset for user 80004 for analysis. The GEO-BLEU scores for each model are shown in Table \ref{Result for User 80004}. The results demonstrate that accuracy significantly improves after adding POI, and approaches near-perfect levels after incorporating time information.

\begin{table}[htbp]
    \centering
    \caption{Result for User 80004 (Single City Normal Scenario)}
    \begin{tabular}{lccc} 
        \toprule 
        Metrics & LP-Bert(Winner of 2023) & Add poi & Add poi time \\
        \midrule 
        GEO-BLEU & 0.568 & 0.934 & 0.948 \\
        \bottomrule 
    \end{tabular}
    \label{Result for User 80004}
\end{table}

\begin{figure}[htbp]
    \centering
    \includegraphics[width=1\linewidth]{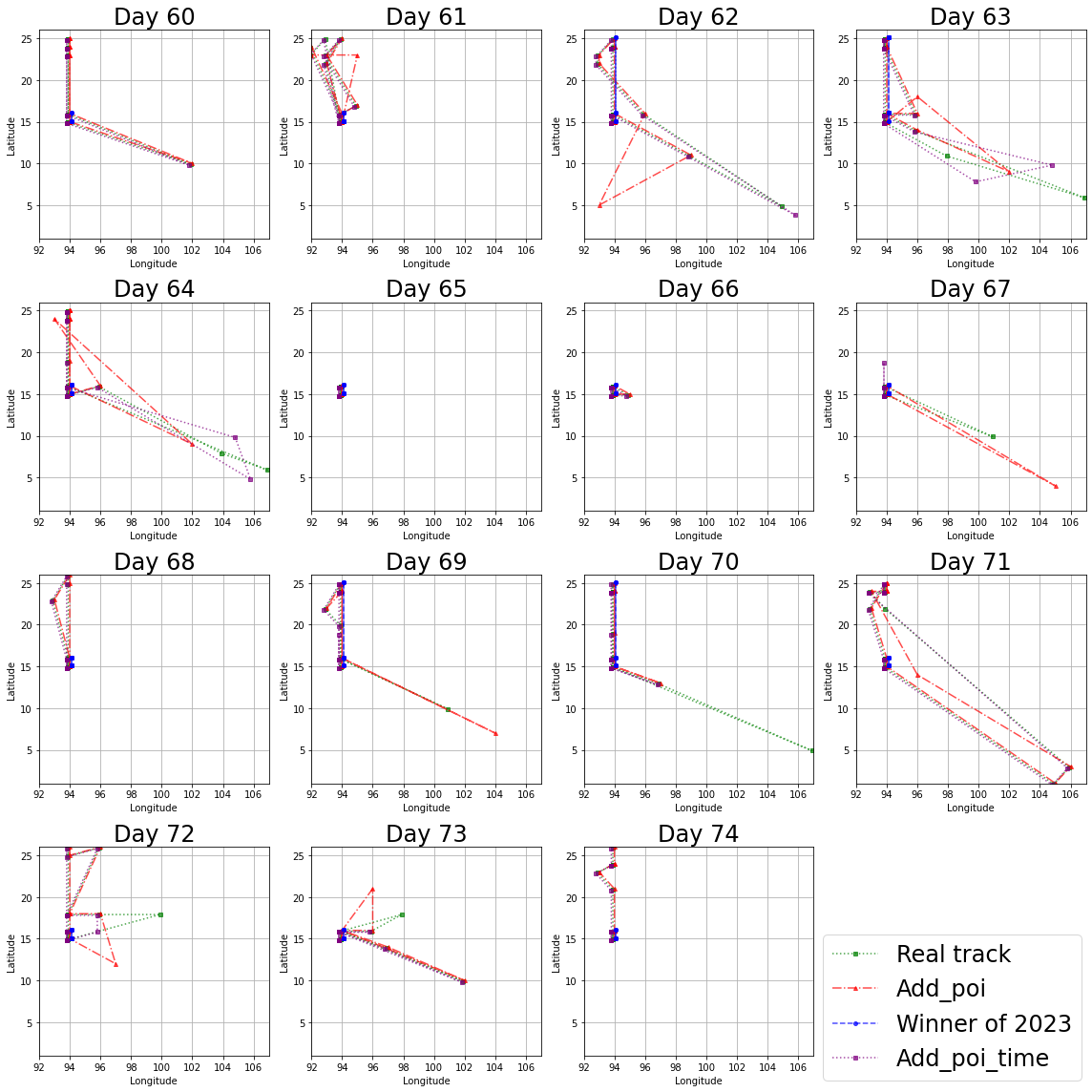}
    \caption{User 80004 Trajectory}
    \label{fig:User 80004 day 71 trajectory}
\end{figure}

Figure \ref{fig:User 80004 day 71 trajectory} shows that the base model predicts daily commutes but misses less regular trips after work. The Add POI model captures both regular commutes and some less regular trips, while the Add POI and Time model offers further refinement, aligning more closely with observed user habits tied to specific days of the week or times.

\label{chap:experiments}

\section{Conclusion}
\label{chap:conclusion}
This paper presents STaBERT, a Semantic-Temporal aware BERT model that effectively integrates both POI embeddings and derived temporal descriptors into a unified deep learning framework for human mobility prediction. By leveraging rich semantic context from points of interest alongside detailed temporal features, STaBERT significantly enhances prediction accuracy in both single-city and multi-city scenarios. Experimental results demonstrate notable improvements, with GEO-BLEU scores increasing from 0.34 to 0.75 for single-city prediction and from 0.34 to 0.56 for multi-city prediction. Despite its architectural simplicity, the model achieves a strong balance between efficiency and performance, offering a practical and scalable solution for real-world urban mobility forecasting tasks. This work highlights the importance of jointly modeling semantic and temporal information to better capture the complexity of human movement patterns in increasingly connected urban environments.
While the main focus of this study is to demonstrate the effectiveness of incorporating POI information, preliminary analysis indicates that in scenarios where POI data is unavailable, prediction performance may decrease. Nevertheless, the embedding module is flexible and can incorporate other spatial descriptors (e.g., land-use, clustering-based location types) in such cases.
\bibliographystyle{ACM-Reference-Format}
\bibliography{References}

%%% -*-BibTeX-*-
%%% Do NOT edit. File created by BibTeX with style
%%% ACM-Reference-Format-Journals [18-Jan-2012].

\begin{thebibliography}{18}

%%% ====================================================================
%%% NOTE TO THE USER: you can override these defaults by providing
%%% customized versions of any of these macros before the \bibliography
%%% command.  Each of them MUST provide its own final punctuation,
%%% except for \shownote{} and \showURL{}.  The latter two
%%% do not use final punctuation, in order to avoid confusing it with
%%% the Web address.
%%%
%%% To suppress output of a particular field, define its macro to expand
%%% to an empty string, or better, \unskip, like this:
%%%
%%% \newcommand{\showURL}[1]{\unskip}   % LaTeX syntax
%%%
%%% \def \showURL #1{\unskip}           % plain TeX syntax
%%%
%%% ====================================================================

\ifx \showCODEN    \undefined \def \showCODEN     #1{\unskip}     \fi
\ifx \showISBNx    \undefined \def \showISBNx     #1{\unskip}     \fi
\ifx \showISBNxiii \undefined \def \showISBNxiii  #1{\unskip}     \fi
\ifx \showISSN     \undefined \def \showISSN      #1{\unskip}     \fi
\ifx \showLCCN     \undefined \def \showLCCN      #1{\unskip}     \fi
\ifx \shownote     \undefined \def \shownote      #1{#1}          \fi
\ifx \showarticletitle \undefined \def \showarticletitle #1{#1}   \fi
\ifx \showURL      \undefined \def \showURL       {\relax}        \fi
% The following commands are used for tagged output and should be
% invisible to TeX
\providecommand\bibfield[2]{#2}
\providecommand\bibinfo[2]{#2}
\providecommand\natexlab[1]{#1}
\providecommand\showeprint[2][]{arXiv:#2}

\bibitem[Amiri et~al\mbox{.}(2024)]%
        {Amiri2024}
\bibfield{author}{\bibinfo{person}{Hossein Amiri}, \bibinfo{person}{Ruochen Kong}, {and} \bibinfo{person}{Andreas Zufle}.} \bibinfo{year}{2024}\natexlab{}.
\newblock \bibinfo{title}{Urban Anomalies: A Simulated Human Mobility Dataset with Injected Anomalies}.
\newblock
\showeprint[arxiv]{2410.01844}~[cs.SI]
\urldef\tempurl%
\url{https://arxiv.org/abs/2410.01844}
\showURL{%
\tempurl}


\bibitem[Devlin et~al\mbox{.}(2019)]%
        {Devlin2018}
\bibfield{author}{\bibinfo{person}{Jacob Devlin}, \bibinfo{person}{Ming-Wei Chang}, \bibinfo{person}{Kenton Lee}, {and} \bibinfo{person}{Kristina Toutanova}.} \bibinfo{year}{2019}\natexlab{}.
\newblock \bibinfo{title}{BERT: Pre-training of Deep Bidirectional Transformers for Language Understanding}.
\newblock
\showeprint[arxiv]{1810.04805}~[cs.CL]
\urldef\tempurl%
\url{https://arxiv.org/abs/1810.04805}
\showURL{%
\tempurl}


\bibitem[Jiang et~al\mbox{.}(2021)]%
        {Jiang2021}
\bibfield{author}{\bibinfo{person}{Renhe Jiang}, \bibinfo{person}{Xuan Song}, \bibinfo{person}{Zipei Fan}, \bibinfo{person}{Tianqi Xia}, \bibinfo{person}{Zhaonan Wang}, \bibinfo{person}{Quanjun Chen}, \bibinfo{person}{Zekun Cai}, {and} \bibinfo{person}{Ryosuke Shibasaki}.} \bibinfo{year}{2021}\natexlab{}.
\newblock \showarticletitle{Transfer Urban Human Mobility via POI Embedding over Multiple Cities}.
\newblock \bibinfo{journal}{\emph{ACM/IMS Trans. Data Sci.}} \bibinfo{volume}{2}, \bibinfo{number}{1}, Article \bibinfo{articleno}{4} (\bibinfo{date}{Jan.} \bibinfo{year}{2021}), \bibinfo{numpages}{26}~pages.
\newblock
\showISSN{2691-1922}
\href{https://doi.org/10.1145/3416914}{doi:\nolinkurl{10.1145/3416914}}


\bibitem[Kim et~al\mbox{.}(2018)]%
        {kim2018}
\bibfield{author}{\bibinfo{person}{Jungmin Kim}, \bibinfo{person}{Juyong Park}, {and} \bibinfo{person}{Wonjae Lee}.} \bibinfo{year}{2018}\natexlab{}.
\newblock \showarticletitle{Why do people move? Enhancing human mobility prediction using local functions based on public records and SNS data}.
\newblock \bibinfo{journal}{\emph{PLOS ONE}} \bibinfo{volume}{13}, \bibinfo{number}{2} (\bibinfo{date}{02} \bibinfo{year}{2018}), \bibinfo{pages}{1--29}.
\newblock
\href{https://doi.org/10.1371/journal.pone.0192698}{doi:\nolinkurl{10.1371/journal.pone.0192698}}


\bibitem[Koyama(2023)]%
        {mobb_koyama23}
\bibfield{author}{\bibinfo{person}{Team~(MOBB) Koyama}.} \bibinfo{year}{2023}\natexlab{}.
\newblock \showarticletitle{MOBB: Ensemble Solution for HuMob 2023}. In \bibinfo{booktitle}{\emph{Proceedings of the HuMob Challenge Workshop at KDD ’23}}. \bibinfo{publisher}{ACM}, \bibinfo{pages}{19--24}.
\newblock


\bibitem[Li et~al\mbox{.}(2023)]%
        {Li23}
\bibfield{author}{\bibinfo{person}{An‐Syu Li}, \bibinfo{person}{Ling‐Huan Meng}, \bibinfo{person}{Yu‐Ling Zhong}, \bibinfo{person}{Yi‐Chung Chen}, {and} \bibinfo{person}{Tomoya Kawakami}.} \bibinfo{year}{2023}\natexlab{}.
\newblock \showarticletitle{LP-BERT: Winning Solution for HuMob 2023}. In \bibinfo{booktitle}{\emph{Proceedings of the HuMob Challenge Workshop at KDD ’23}}. \bibinfo{publisher}{ACM}, \bibinfo{pages}{7--12}.
\newblock


\bibitem[Ratti et~al\mbox{.}(2006)]%
        {Ratti2006}
\bibfield{author}{\bibinfo{person}{Carlo Ratti}, \bibinfo{person}{Dennis Frenchman}, \bibinfo{person}{Riccardo~Maria Pulselli}, {and} \bibinfo{person}{Sarah Williams}.} \bibinfo{year}{2006}\natexlab{}.
\newblock \showarticletitle{Mobile Landscapes: Using Location Data from Cell Phones for Urban Analysis}.
\newblock \bibinfo{journal}{\emph{Environment and Planning B: Planning and Design}} \bibinfo{volume}{33}, \bibinfo{number}{5} (\bibinfo{year}{2006}), \bibinfo{pages}{727--748}.
\newblock
\showeprint{https://doi.org/10.1068/b32047}
\href{https://doi.org/10.1068/b32047}{doi:\nolinkurl{10.1068/b32047}}


\bibitem[Shimizu et~al\mbox{.}(2022)]%
        {Shimizu2022}
\bibfield{author}{\bibinfo{person}{Toru Shimizu}, \bibinfo{person}{Kota Tsubouchi}, {and} \bibinfo{person}{Takahiro Yabe}.} \bibinfo{year}{2022}\natexlab{}.
\newblock \showarticletitle{GEO-BLEU: Similarity Measure for Geospatial Sequences}. In \bibinfo{booktitle}{\emph{Proceedings of the 30th ACM SIGSPATIAL International Conference on Advances in Geographic Information Systems (SIGSPATIAL ’22)}} (Seattle, WA, USA). \bibinfo{publisher}{ACM}, \bibinfo{pages}{1--4}.
\newblock
\href{https://doi.org/10.1145/3557915.3560951}{doi:\nolinkurl{10.1145/3557915.3560951}}


\bibitem[Solatorio(2023)]%
        {Solatorio2023}
\bibfield{author}{\bibinfo{person}{Aivin~V. Solatorio}.} \bibinfo{year}{2023}\natexlab{}.
\newblock \showarticletitle{GeoFormer: Predicting Human Mobility using Generative Pre-trained Transformer (GPT)}. In \bibinfo{booktitle}{\emph{Proceedings of the 1st International Workshop on the Human Mobility Prediction Challenge}} \emph{(\bibinfo{series}{HuMob-Challenge ’23})}. \bibinfo{publisher}{ACM}, \bibinfo{pages}{11–15}.
\newblock
\href{https://doi.org/10.1145/3615894.3628499}{doi:\nolinkurl{10.1145/3615894.3628499}}


\bibitem[Suzuki et~al\mbox{.}(2024)]%
        {Meisaku2024}
\bibfield{author}{\bibinfo{person}{Meisaku Suzuki}, \bibinfo{person}{Yusuke Fukushima}, \bibinfo{person}{Ryo Koyama}, \bibinfo{person}{Hayato Kumagai}, \bibinfo{person}{Tomohiro Mimura}, {and} \bibinfo{person}{Keiichi Ochiai}.} \bibinfo{year}{2024}\natexlab{}.
\newblock \showarticletitle{Cross-city-aware Spatiotemporal BERT}. In \bibinfo{booktitle}{\emph{Proceedings of the 2nd ACM SIGSPATIAL International Workshop on Human Mobility Prediction Challenge}} (Atlanta, GA, USA) \emph{(\bibinfo{series}{HuMob'24})}. \bibinfo{publisher}{Association for Computing Machinery}, \bibinfo{address}{New York, NY, USA}, \bibinfo{pages}{33–36}.
\newblock
\showISBNx{9798400711503}
\href{https://doi.org/10.1145/3681771.3699915}{doi:\nolinkurl{10.1145/3681771.3699915}}


\bibitem[Suzuki et~al\mbox{.}(2023a)]%
        {Suzuki2023}
\bibfield{author}{\bibinfo{person}{Masahiro Suzuki}, \bibinfo{person}{Shomu Furuta}, {and} \bibinfo{person}{Yusuke Fukazawa}.} \bibinfo{year}{2023}\natexlab{a}.
\newblock \showarticletitle{Personalized human mobility prediction for HuMob challenge}. In \bibinfo{booktitle}{\emph{Proceedings of the 1st International Workshop on the Human Mobility Prediction Challenge}} (Hamburg, Germany) \emph{(\bibinfo{series}{HuMob-Challenge '23})}. \bibinfo{publisher}{Association for Computing Machinery}, \bibinfo{address}{New York, NY, USA}, \bibinfo{pages}{22–25}.
\newblock
\showISBNx{9798400703560}
\href{https://doi.org/10.1145/3615894.3628501}{doi:\nolinkurl{10.1145/3615894.3628501}}


\bibitem[Suzuki et~al\mbox{.}(2023b)]%
        {Masahiro2023}
\bibfield{author}{\bibinfo{person}{Masahiro Suzuki}, \bibinfo{person}{Shomu Furuta}, {and} \bibinfo{person}{Yusuke Fukazawa}.} \bibinfo{year}{2023}\natexlab{b}.
\newblock \showarticletitle{Personalized human mobility prediction for HuMob challenge}. In \bibinfo{booktitle}{\emph{Proceedings of the 1st International Workshop on the Human Mobility Prediction Challenge}} (Hamburg, Germany) \emph{(\bibinfo{series}{HuMob-Challenge '23})}. \bibinfo{publisher}{Association for Computing Machinery}, \bibinfo{address}{New York, NY, USA}, \bibinfo{pages}{22–25}.
\newblock
\showISBNx{9798400703560}
\href{https://doi.org/10.1145/3615894.3628501}{doi:\nolinkurl{10.1145/3615894.3628501}}


\bibitem[Tang et~al\mbox{.}(2024)]%
        {Llama-3-8b}
\bibfield{author}{\bibinfo{person}{Peizhi Tang}, \bibinfo{person}{Chuang Yang}, \bibinfo{person}{Tong Xing}, \bibinfo{person}{Xiaohang Xu}, \bibinfo{person}{Renhe Jiang}, {and} \bibinfo{person}{Kaoru Sezaki}.} \bibinfo{year}{2024}\natexlab{}.
\newblock \showarticletitle{Instruction-Tuning Llama-3-8B Excels in City-Scale Mobility Prediction}. In \bibinfo{booktitle}{\emph{Proceedings of the 2nd ACM SIGSPATIAL International Workshop on Human Mobility Prediction Challenge}} (Atlanta, GA, USA) \emph{(\bibinfo{series}{HuMob'24})}. \bibinfo{publisher}{Association for Computing Machinery}, \bibinfo{address}{New York, NY, USA}, \bibinfo{pages}{1–4}.
\newblock
\showISBNx{9798400711503}
\href{https://doi.org/10.1145/3681771.3699908}{doi:\nolinkurl{10.1145/3681771.3699908}}


\bibitem[Terashima et~al\mbox{.}(2024)]%
        {Multiple_BERTs}
\bibfield{author}{\bibinfo{person}{Haru Terashima}, \bibinfo{person}{Shun Takagi}, \bibinfo{person}{Naoki Tamura}, \bibinfo{person}{Kazuyuki Shoji}, \bibinfo{person}{Tahera Hossain}, \bibinfo{person}{Shin Katayama}, \bibinfo{person}{Kenta Urano}, \bibinfo{person}{Takuro Yonezawa}, {and} \bibinfo{person}{Nobuo Kawaguchi}.} \bibinfo{year}{2024}\natexlab{}.
\newblock \showarticletitle{Time-series Stay Frequency for Multi-City Next Location Prediction using Multiple BERTs} \emph{(\bibinfo{series}{HuMob'24})}. \bibinfo{publisher}{Association for Computing Machinery}, \bibinfo{address}{New York, NY, USA}, \bibinfo{pages}{5–9}.
\newblock
\showISBNx{9798400711503}
\href{https://doi.org/10.1145/3681771.3699909}{doi:\nolinkurl{10.1145/3681771.3699909}}


\bibitem[Terashima et~al\mbox{.}(2023)]%
        {Terashima2023}
\bibfield{author}{\bibinfo{person}{Haru Terashima}, \bibinfo{person}{Naoki Tamura}, \bibinfo{person}{Kazuyuki Shoji}, \bibinfo{person}{Shin Katayama}, \bibinfo{person}{Kenta Urano}, \bibinfo{person}{Takuro Yonezawa}, {and} \bibinfo{person}{Nobuo Kawaguchi}.} \bibinfo{year}{2023}\natexlab{}.
\newblock \showarticletitle{Human Mobility Prediction Challenge: Next Location Prediction using Spatiotemporal BERT}. In \bibinfo{booktitle}{\emph{Proceedings of the 1st International Workshop on the Human Mobility Prediction Challenge}} (Hamburg, Germany) \emph{(\bibinfo{series}{HuMob-Challenge '23})}. \bibinfo{publisher}{Association for Computing Machinery}, \bibinfo{address}{New York, NY, USA}, \bibinfo{pages}{1–6}.
\newblock
\showISBNx{9798400703560}
\href{https://doi.org/10.1145/3615894.3628498}{doi:\nolinkurl{10.1145/3615894.3628498}}


\bibitem[Vaswani et~al\mbox{.}(2017)]%
        {vaswani2017}
\bibfield{author}{\bibinfo{person}{Ashish Vaswani}, \bibinfo{person}{Noam Shazeer}, \bibinfo{person}{Niki Parmar}, \bibinfo{person}{Jakob Uszkoreit}, \bibinfo{person}{Llion Jones}, \bibinfo{person}{Aidan~N Gomez}, \bibinfo{person}{{\L}ukasz Kaiser}, {and} \bibinfo{person}{Illia Polosukhin}.} \bibinfo{year}{2017}\natexlab{}.
\newblock \showarticletitle{Attention is all you need}.
\newblock \bibinfo{journal}{\emph{Advances in neural information processing systems}}  \bibinfo{volume}{30} (\bibinfo{year}{2017}).
\newblock


\bibitem[Wang et~al\mbox{.}(2019)]%
        {wang2019}
\bibfield{author}{\bibinfo{person}{Jinzhong Wang}, \bibinfo{person}{Xiangjie Kong}, \bibinfo{person}{Feng Xia}, {and} \bibinfo{person}{Lijun Sun}.} \bibinfo{year}{2019}\natexlab{}.
\newblock \showarticletitle{Urban Human Mobility: Data-Driven Modeling and Prediction}.
\newblock \bibinfo{journal}{\emph{SIGKDD Explor. Newsl.}} \bibinfo{volume}{21}, \bibinfo{number}{1} (\bibinfo{date}{May} \bibinfo{year}{2019}), \bibinfo{pages}{1–19}.
\newblock
\showISSN{1931-0145}
\href{https://doi.org/10.1145/3331651.3331653}{doi:\nolinkurl{10.1145/3331651.3331653}}


\bibitem[Wikipedia(6 05)]%
        {dtw_wiki2025}
\bibfield{author}{\bibinfo{person}{Wikipedia}.} \bibinfo{year}{Accessed 2025-06-05}\natexlab{}.
\newblock \bibinfo{title}{Dynamic Time Warping}.
\newblock \bibinfo{howpublished}{\url{https://en.wikipedia.org/wiki/Dynamic_time_warping}}.
\newblock
\newblock
\shownote{Online encyclopedia entry describing the DTW algorithm}.


\end{thebibliography}

\end{document}